\documentclass{article}

\usepackage{ijcai13}

\usepackage{times}
\usepackage{helvet}
\usepackage{courier}

\usepackage{amsmath}
\usepackage{amssymb}

\usepackage{latexsym}
\usepackage{picinpar}

\newtheorem {definition}{Definition}
\newtheorem {proposition}{Proposition}

\newtheorem {theorem}{Theorem}

\newtheorem {example}{Example}

\begin{document}

\title{Fuzzy Aggregates in Fuzzy Answer Set Programming}
\author{ Emad Saad \\
emsaad@gmail.com
}

\maketitle

\begin{abstract}
Fuzzy answer set programming \cite{Saad_DFLP,Saad_EFLP,Subrahmanian_B} is a declarative framework for representing and reasoning about knowledge in fuzzy environments. However, the unavailability of fuzzy aggregates in disjunctive fuzzy logic programs, DFLP, with fuzzy answer set semantics \cite{Saad_DFLP} prohibits the natural and concise representation of many interesting problems. In this paper, we extend DFLP to allow arbitrary fuzzy aggregates. We define fuzzy answer set semantics for DFLP with arbitrary fuzzy aggregates including monotone, antimonotone, and nonmonotone fuzzy aggregates. We show that the proposed fuzzy answer set semantics subsumes both the original fuzzy answer set semantics of DFLP \cite{Saad_DFLP} and the classical answer set semantics of classical disjunctive logic programs with classical aggregates \cite{Recur-aggr}, and consequently subsumes the classical answer set semantics of classical disjunctive logic programs \cite{Gelfond_B}. We show that the proposed fuzzy answer sets of DFLP with fuzzy aggregates are minimal fuzzy models and hence incomparable, which is an important property for nonmonotonic fuzzy reasoning.
\end{abstract}

\section{Introduction}

Fuzzy answer set programming \cite{Saad_DFLP,Saad_EFLP,Subrahmanian_B} is a declarative programming framework that has been shown effective for knowledge representation and reasoning in fuzzy  environments. These include representing and reasoning about actions with fuzzy effects and fuzzy planning \cite{Saad_EFLP,Saad_IFSA} as well as representing and reasoning about fuzzy preferences \cite{Saad_DFLP}. The fuzzy answer set programming framework includes disjunctive fuzzy logic programs \cite{Saad_DFLP}, extended fuzzy logic programs \cite{Saad_EFLP}, and normal fuzzy logic programs \cite{Subrahmanian_B} with fuzzy answer set semantics. However, the unavailability of fuzzy aggregates in fuzzy answer set programming \cite{Saad_DFLP,Saad_EFLP,Subrahmanian_B} disallows the natural and concise representation of many new interesting problems.

\begin{example} Consider the same company control problem described in \cite{Recur-aggr}. Assume that a company $C_1$ owns $P \: \%$ of a company $C_2$ shares, represented by the predicate $ownsStk(C_1, C_2, P)$. If the company $C_1$ owns a total sum of more than $50 \: \%$ of shares of the company $C_2$ directly (through $C_1$ itself) or indirectly (through another company $C_3$ controlled by $C_1$), then we say that company $C_1$ controls company $C_2$. Let $controls(C_1, C_2)$ denotes that company $C_1$ controls company $C_2$. Let $controlStk(C_1, C_2, C_3, P)$ denotes that company $C_1$ controls $P \: \%$ of company $C_3$ shares through company $C_2$, since $C_1$ controls $C_2$ and $C_2$ owns $P \: \%$ of $C_3$ shares. Assume information about companies shares are represented as facts as described below. This company control problem is represented as a classical disjunctive logic program with classical aggregates, described below, whose answer set describes the intuitive and correct solution to the problem as illustrated in \cite{Recur-aggr} as:
\[
\begin{array}{lcl}
ownsStk(a, b, 40) & \leftarrow &
\\
ownsStk(c, b, 20) & \leftarrow &
\\
ownsStk(a, c, 40) & \leftarrow &
\\
ownsStk(b, c, 20) & \leftarrow &
\\
controlStk(C_1, C_1, C_2, P) & \leftarrow & ownsStk(C_1, C_2, P).
\\
controlStk(C_1, C_2, C_3, P) & \leftarrow & controls(C_1, C_2), \\ && ownsStk(C_2, C_3, P).
\end{array}
\]
\[
\begin{array}{l}
controls(C_1, C_3)  \leftarrow  \\ \qquad \qquad sum \{P, C_2, controlStk(C_1, C_2, C_3, P) \} > 50.
\end{array}
\]
\label{ex:company}
\end{example}
The above representation of the company control problem as a classical disjunctive logic program with classical aggregates is entirely correct if our knowledge regarding the companies shares are prefect. However this is not always the case. Consider our knowledge regarding the company shares is not perfect. Thus, we cannot absolutely assert that some company $C_1$ controls another company $C_2$ as in the above representation. Instead, we can assert that a company $C_1$ controls another company $C_2$ with a certain degree of beliefs. In the presence of such uncertainties, the above company control problem need to be redefined to deal with imperfect knowledge about companies shares (namely fuzzy company control problem), where the imperfect knowledge about the companies shares are represented as a {\em fuzzy set} over companies shares. Consequently, a logical framework different from classical disjunctive logic programs with classical aggregates is needed for representing and reasoning about such fuzzy reasoning problems.

Consider that the fuzzy set over companies shares, presented in Example (\ref{ex:company}), is described as; company $a$ owns $40 \: \%$ of company $b$ with grade membership $0.7$; company $c$ owns $20 \: \%$ of company $b$ with grade membership $0.6$; company $a$ owns $40 \: \%$ of company $c$ with grade membership $0.9$; and company $b$ owns $20 \: \%$ of company $c$ with grade membership $0.8$. Consider also that the same company control strategy as in Example (\ref{ex:company}) is employed. Thus, this fuzzy company control problem cannot be represented as a classical disjunctive logic program with classical aggregates, since classical disjunctive logic programs with classical aggregates do not allow neither representing and reasoning in the presence of fuzzy uncertainty nor allow aggregation over fuzzy sets. Moreover, this fuzzy company control problem cannot be represented as a disjunctive fuzzy logic program with fuzzy answer set semantics either, since disjunctive fuzzy logic programs with fuzzy answer set semantics do not allow aggregations over fuzzy sets by means of {\em fuzzy aggregates} for intuitive and concise representation of the problem.

Therefore, we propose to extend disjunctive fuzzy logic programs with fuzzy answer set semantics \cite{Saad_DFLP}, denoted by DFLP, with arbitrary fuzzy aggregates to allow intuitive and concise representation of many real-world problems. To the best of our knowledge, this development is the first that defines semantics for fuzzy aggregates in a fuzzy answer set programming framework.

The contributions of this paper are as follows. We extend the original language of DFLP to allow arbitrary fuzzy annotation function including monotone, antimonotone, and nonmonotone annotation functions. We define the notions of fuzzy aggregates and fuzzy aggregate atoms in DFLP. We develop the fuzzy answer set semantics of DFLP with arbitrary fuzzy aggregates, denoted by DFLP$^{\cal FA}$, including monotone, antimonotone, and nonmonotone fuzzy aggregates. We show that the presented fuzzy answer set semantics of DFLP$^{\cal FA}$ subsumes and generalizes both the original fuzzy answer set semantics of DFLP \cite{Saad_DFLP} and the classical answer set semantics of the classical disjunctive logic programs with classical aggregates, denoted by DLP$^{\cal A}$ \cite{Recur-aggr}, and consequently subsumes the classical answer set semantics of classical disjunctive logic programs, denoted by DLP \cite{Gelfond_B}. We show that the fuzzy answer sets of DFLP$^{\cal FA}$ are minimal fuzzy models and hence incomparable, which is an important property for nonmonotonic fuzzy reasoning.

The choice of DFLP for extension with fuzzy aggregates is interesting for many reasons. First, DFLP is very expressive form of fuzzy answer set programming that allows disjunctions to appear in the head of rules. It has been shown in \cite{Saad_DFLP} that; (1) DFLP is capable of representing and reasoning with both fuzzy uncertainty and qualitative uncertainty in which fuzzy uncertainly need to be defined over qualitative uncertainty; (2) DFLP is shown to be sophisticated logical framework for representing and reasoning about fuzzy preferences; (3) DFLP is a natural extension to DLP and its fuzzy answer set semantics subsumes the classical answer set semantics of DLP \cite{Gelfond_B}; (4) DFLP with fuzzy answer set semantics subsumes the fuzzy answer set programming framework of \cite{Subrahmanian_B}, which are DFLP programs with only an atom appearing in heads of rules.

\section{DFLP$^{\cal FA}$ : Fuzzy Aggregates Disjunctive Fuzzy Logic Programs}

In this section we present the basic language of DFLP$^{\cal FA}$, the notions of fuzzy aggregates and fuzzy aggregate atoms, and the syntax of DFLP$^{\cal FA}$ programs.

\subsection{The Basic Language of DFLP$^{\cal FA}$}

Let $\cal L$ denotes an arbitrary first-order language with finitely many predicate symbols, function symbols, constants, and infinitely many variables. A term is a constant, a variable or a function. An atom, $a$, is a predicate in $\cal L$, where $\cal {B_L}$ is the Herbrand base of $\cal L$. The Herbrand universe of $\cal L$ is denoted by $U_{\cal L}$. Non-monotonic negation or the negation as failure is denoted by $not$. In fuzzy aggregates disjunctive fuzzy logic programs, DFLP$^{\cal FA}$, the grade membership values are assigned to atoms in $\cal {B_L}$ as values from $[0,1]$. The set $[0,1]$ and the relation $\leq$ form a complete lattice, where the join ($\oplus$) operation is defined as $\alpha_1 \oplus \alpha_2 = \max (\alpha_1,\alpha_2)$ and the meet ($\otimes$) is defined as $\alpha_1 \otimes \alpha_2 = \min (\alpha_1,\alpha_2)$.

A \emph{fuzzy annotation}, $\mu$, is either a constant in $[0, 1]$ (called \emph{fuzzy annotation constant}), a variable ranging over $[0, 1]$ (called \emph{fuzzy annotation variable}), or $f(\alpha_1,\ldots,\,\alpha_n)$ (called \emph{fuzzy annotation function}) where $f$ is a representation of a monotone, antimonotone, or nonmonotone total or partial function $f: ([0, 1])^n \rightarrow [0, 1]$ and $\alpha_1,\ldots,\alpha_n$ are fuzzy annotations. If $a$ is an atom and $\mu$ is a fuzzy annotation then $a:\mu$ is called a fuzzy annotated atom.

\subsection{Fuzzy Aggregate Atoms}

A symbolic fuzzy set is an expression of the form \\ $\{ X : U \; | \; C \}$, where $X$ is a variable or a function term and $U$ is fuzzy annotation variable or fuzzy annotation function, and $C$ is a conjunction of fuzzy annotated atoms. A ground fuzzy set is a set of pairs of the form $\langle X^g : U^g \; | \; C^g \rangle$ such that $X^g$ is a constant term and $U^g$ is fuzzy annotation constant, and $C^g$ is a ground conjunction of fuzzy annotated atoms. A symbolic fuzzy set or ground fuzzy set is called a fuzzy set term. Let $f$ be a fuzzy aggregate function symbol and $S$ be a fuzzy set term, then $f(S)$ is said a fuzzy aggregate, where $f \in \{$$sum_F$, $times_F$, $min_F$, $max_F$, $count_F$$\}$. If $f(S)$ is a fuzzy aggregate and $T$ is a constant, a variable or a function term, called {\em guard}, then we say $f(S) \prec T$ is a fuzzy aggregate atom, where $\prec \in \{=, \neq, <, >, \leq, \geq \}$.

\begin{example} The following are examples for fuzzy aggregate atoms representation in DFLP$^{\cal FA}$ language.
\[
\begin{array}{c}
max_F \{ X : U \: | \: benefit(X): U  \} >  99 \\
sum_F \{ \langle 2: 0.4 \:|\:  \: a(1, 2): 0.4 \rangle, \; \langle 5:0.7 \:|\: \: a(1, 5):0.7 \rangle \} \leq  11 \\
\end{array}
\]
\label{ex:samples}
\end{example}
\begin{definition} Let $f(S)$ be a fuzzy aggregate. A variable, $X$, is a local variable to $f(S)$ if and only if $X$ appears in $S$ and $X$ does not appear in the DFLP$^{\cal FA}$ rule that contains $f(S)$.
\label{def:local}
\end{definition}
Definition (\ref{def:local}) characterizes the local variables for a fuzzy aggregate function. For example, for the first fuzzy aggregate atom in Example (\ref{ex:samples}), the variables $X$ and $U$ are local variables to the fuzzy aggregate $max_F$.
\begin{definition} A global variable is a variable that is not a local variable.
\end{definition}

\subsection{DFLP$^{\cal FA}$ Program Syntax}

This section defines the syntax of rules and programs in the language of DFLP$^{\cal FA}$.

\begin{definition} A DFLP$^{\cal FA}$ rule is an expression of the form
\[
\begin{array}{r}
a_1:\mu_1 \; \vee \ldots \vee \; a_k:\mu_k \leftarrow a_{k+1}:\mu_{k+1}, \ldots, a_m:\mu_m, \\ not\; a_{m+1}:\mu_{m+1},\ldots, not\;a_{n}:\mu_{n},
\end{array}
\]
where $\forall (1 \leq i \leq k)$ $a_i$ are atoms, $\forall (k+1 \leq i \leq n)$ $a_i$ are atoms or fuzzy aggregate atoms, and $\forall (1 \leq i \leq n)$ $\mu_i$  are fuzzy annotations.
\end{definition}
A DFLP$^{\cal FA}$ rule means that if for each $a_i:\mu_i$, where $k+1 \leq i \leq m$, it is {\em believable} that the grade membership value of $a_i$ is at least $\mu_i$ w.r.t. $\leq$ and for each $not\; a_j:\mu_j$, where $m+1 \leq j \leq n$, it is \emph{not believable} that the grade membership value of $a_j$ is at least $\mu_j$ w.r.t. $\leq$, then there exists at least $a_i$, where $1 \leq i \leq k$, such that the grade membership value of $a_i$ is at least $\mu_i$.
\begin{definition}
A DFLP$^{\cal FA}$ program, $\Pi$, is a set of DFLP$^{\cal FA}$ rules.
\end{definition}
For the simplicity of the presentation, atoms that appear in DFLP$^{\cal FA}$ programs without fuzzy annotations are assumed to be associated with the fuzzy annotation constant $1$.

\begin{example} The fuzzy company control problem described in Example \ref{ex:company} can be concisely and intuitively represented as DFLP$^{\cal FA}$ program, $\Pi$, that consists of the DFLP$^{\cal FA}$ rules:
\[
\begin{array}{lcl}
ownsStk(a, b, 40):0.7 & \leftarrow &
\\
ownsStk(c, b, 20):0.6 &  \leftarrow &
\\
ownsStk(a, c, 40):0.9 & \leftarrow &
\\
ownsStk(b, c, 20):0.8 & \leftarrow &
\end{array}
\]

\[
\begin{array}{l}
controlStk(C_1, C_1, C_2, P):V  \leftarrow  \\  \qquad \qquad \qquad \qquad \qquad \qquad ownsStk(C_1, C_2, P):V
\\
controlStk(C_1, C_2, C_3, P): V  \leftarrow  \\  \qquad \quad controls(C_1, C_2):0.55, ownsStk(C_2, C_3, P):V \\
\end{array}
\]
\[
\begin{array}{l}
controls(C_1, C_3): 0.55 \leftarrow \\ sum_F \{P : V \; | \; controlStk(C_1, C_2, C_3, P):V \} > 50 \; : \; 0.6
\end{array}
\]
The last DFLP$^{\cal FA}$ rule in, $\Pi$, says that if it is at least $0.6$ grade membership value believable that company $C_1$ owns a total sum of more than $50 \: \%$ of shares of the company $C_3$ directly (through $C_1$ itself) or indirectly (through another company $C_2$ controlled by $C_1$), then it is $0.55$ grade membership value believable that company $C_1$ controls company $C_3$.
\label{ex:main}
\end{example}

\begin{definition}
The {\em ground instantiation} of a symbolic fuzzy set $$S = \{ X : U  \; | \; C \}$$ is the set of all ground pairs of the form $\langle \theta\; (X) : \theta\; (U) \; | \;  \theta \; (C) \rangle$, where $\theta$ is a substitution of every local variable appearing in $S$ to a constant from $U_{\cal L}$.
\end{definition}

\begin{definition} A ground instantiation of a DFLP$^{\cal FA}$ rule, $r$, is the replacement of each global variable appearing in $r$ to a constant from $U_{\cal L}$, then followed by the ground instantiation of every symbolic fuzzy set, $S$, appearing in $r$.

The ground instantiation of a DFLP$^{\cal FA}$ program, $\Pi$, is the set of all possible ground instantiations of every DFLP$^{\cal FA}$ rule in $\Pi$.
\end{definition}

\begin{example} A ground instantiation of the DFLP$^{\cal FA}$ rule
\[
\begin{array}{l}
controls(C_1, C_3): 0.55  \leftarrow \\ sum_F \{P : V \; | \; controlStk(C_1, C_2, C_3, P):V \} > 50 \; : \; 0.6
\end{array}
\]
with respect to the DFLP$^{\cal FA}$ program, $\Pi$, described in Example \ref{ex:main}, is given as:
\[
\begin{array}{lcl}
controls(a, c) && : 0.55 \leftarrow   sum_F \{ \\ &&
\langle  40:0.9 \:|\:  controlStk(a, a, c, 40):0.9  \rangle, \\ &&
\langle  0:0.0  \:|\:  controlStk(a, b, c, 0):0.0  \rangle, \\ &&
\langle  0:0.0  \:|\:  controlStk(a, c, c, 0):0.0  \rangle, \\ &&
\ldots \} \; > \; 50 \; : \; 0.6
\end{array}
\]
\end{example}

\section {Fuzzy Aggregates Semantics}

A fuzzy aggregate is an aggregation over a fuzzy set that returns the evaluation of a classical aggregate and the grade membership value of the evaluation of that classical aggregate over a given fuzzy set. The fuzzy aggregates that we consider are $sum_F$, $times_F$, $min_F$, $max_F$, and $count_F$ that find the evaluation of the classical aggregates $sum$, $times$, $min$, $max$, and $count$ respectively along with the grade membership value of their evaluations. The application of fuzzy aggregates is on ground fuzzy sets which are sets of constants terms along with their associated grade membership values.

\subsection{Mappings}

Let $\mathbb{X}$ denotes a set of objects. Then, we use $2^\mathbb{X}$ to denote the set of all {\em multisets} over elements in $\mathbb{X}$. Let $\mathbb{R}$ denotes the set of all real numbers and $\mathbb{N}$ denotes the set of all natural numbers, and $U_{\cal L}$ denotes the Herbrand universe. Let $\bot$ be a symbol that does not occur in ${\cal L}$. Therefore, the mappings of the fuzzy aggregates are given by:

\begin{itemize}

\item $sum_F : 2^{\mathbb{R} \times [0, 1] } \rightarrow \mathbb{R} \times [0, 1]$.

\item $times_F: 2^{\mathbb{R} \times [0, 1] } \rightarrow \mathbb{R} \times [0, 1]$.

\item $min_F : (2^{\mathbb{R} \times [0, 1] } - \emptyset) \rightarrow
    \mathbb{R} \times [0, 1]$.

\item $max_F: (2^{\mathbb{R} \times [0, 1] } - \emptyset) \rightarrow
    \mathbb{R} \times [0, 1]$.

\item $count_F : 2^{U_{\cal L} \times [0, 1]}  \rightarrow \mathbb{N} \times [0, 1]$.

\end{itemize}
The application of $sum_F$ and $times_F$ on the empty multiset return $(0,1)$ and $(1,1)$ respectively. The application of $count_F$ on the empty multiset returns $(0,1)$. However, the application of $max_F$ and $min_F$  on the empty multiset is undefined.

\begin{definition}
A fuzzy interpretation of a DFLP$^{\cal FA}$ program, $\Pi$, is a mapping $I:{\cal B_L} \rightarrow [0, 1]$.
\end{definition}

\subsection{Semantics of Fuzzy Aggregates}

The semantics of fuzzy aggregates is defined with respect to a fuzzy interpretation, which is a representation of fuzzy sets. A fuzzy annotated atom, $a:\mu$, is true (satisfied) with respect to a fuzzy interpretation, $I$, if and only if $\mu \leq I(a)$. The negation of a fuzzy annotated atom, $not \; a:\mu$, is true (satisfied) with respect to $I$ if and only if $\mu \nleq I(a)$. The evaluation of a fuzzy aggregate, and hence the truth valuation of a fuzzy aggregate atom, are established with respect to a given fuzzy interpretation, $I$, as presented in the following definitions.

\begin{definition} Let $f(S)$ be a ground fuzzy aggregate and $I$ be a fuzzy interpretation. Then, we define $S_I$ to be the multiset constructed from elements in $S$, where $S_I = \{\!\!\{ X^g : U^g  \; | \; \\ \langle X^g : U^g \; | \; C^g \rangle \in S \wedge$ $C^g$ is true w.r.t. $I \}\!\!\}$.
\end{definition}

\begin{definition} Let $f(S)$ be a ground fuzzy aggregate and $I$ be a fuzzy interpretation. Then, the evaluation of $f(S)$ with respect to $I$ is, $f(S_I)$, the result of the application of $f$ to $S_I$, where $f(S_I) = \bot$ if $S_I$ is not in the domain of $f$ and

\begin{itemize}

\item $sum_F(S_I) = (\sum_{\: X^g : U^g \in S_I} \; X^g \; , \;  \min_{\: X^g : U^g  \in S_I} \; U^g ) $

\item $times_F(S_I) = (\prod_{\: X^g : U^g \in S_I} \; X^g \; , \;  \min_{\: X^g : U^g  \in S_I} \; U^g ) $

\item $min_F (S_I)= (\min_{\: X^g : U^g \in S_I} \; X^g \; , \;  \min_{\: X^g : U^g  \in S_I} \; U^g ) $

\item $max_F (S_I)= (\max_{\: X^g : U^g \in S_I} \; X^g \; , \;  \min_{\: X^g : U^g  \in S_I} \; U^g ) $

\item $count_F (S_I)= (count_{\: X^g : U^g \in S_I} \; X^g \; , \;  \min_{\: X^g : U^g  \in S_I} \; U^g )$

\end{itemize}
\label{def:ExpProb}
\end{definition}

\section{Fuzzy Answer Set Semantics of DFLP$^{\cal FA}$}

In this section we define the satisfaction, fuzzy models, and the fuzzy answer set semantics of fuzzy aggregates disjunctive fuzzy logic programs, DFLP$^{\cal FA}$. Let $r$ be a DFLP$^{\cal FA}$ rule and $head(r) = a_1:\mu_1 \; \vee \ldots \vee \; a_k:\mu_k$ and \\ $body(r) = a_{k+1}:\mu_{k+1}, \ldots, a_m:\mu_m, not\;a_{m+1}:\mu_{m+1},\ldots, not\;a_{n}:\mu_{n}$.

\begin{definition} Let $\Pi$ be a ground DFLP$^{\cal FA}$ program, $r$ be a DFLP$^{\cal FA}$ rule in $\Pi$, $I$ be a fuzzy interpretation for $\Pi$, and $f \in \{sum_F, times_F, min_F , max_F, count_F \}$. Then,

\begin{enumerate}

\item $I$ satisfies $a_i:\mu_i$ in $head(r)$ iff  $\mu_i \leq I(a_i)$.

\item $I$ satisfies $f(S) \prec T : \mu$ in $body (r)$ iff $f(S_I) = (x, \nu) \neq \bot$ and $x \prec T$ and $\mu \leq \nu$.

\item $I$ satisfies $not \; f(S) \prec T :\mu $ in $body (r)$ iff $f(S_I) =  \bot$ or $f(S_I) = (x, \nu) \neq \bot$ and $x \nprec T$ or $\mu \nleq \nu$.

\item $I$ satisfies $a_i : \mu_i$ in $body(r)$ iff  $\mu_i \leq I(a_i)$.

\item $I$ satisfies $not\;a_j:\mu_j$ in $body(r)$ iff $\mu_j \nleq I(a_j)$.

\item $I$ satisfies $body(r)$ iff $\forall(k+1 \leq i \leq m), I$ satisfies $a_i : \mu_i$ and $\forall(m+1 \leq j
\leq n), I$ satisfies $not\;(a_j : \mu_j)$.

\item $I$ satisfies $head(r)$ iff $\exists i$ $(1 \leq i \leq k)$ such that $I$ satisfies $a_i : \mu_i$.

\item $I$ satisfies $r$ iff $I$ satisfies $head(r)$ whenever $I$ satisfies $body(r)$ or $I$ does not satisfy $body(r)$.

\item $I$ satisfies $\Pi$ iff $I$ satisfies every DFLP$^{\cal FA}$ rule in $\Pi$ and

\begin{itemize}
\item $\max \{\!\!\{\mu_i \;  \; | \; head(r) \leftarrow body(r) \in \Pi \}\!\!\}\leq I(a_i)$ such that $I$ satisfies $body(r)$ and $I$ satisfies $a_i:\mu_i$ in the $head(r)$.

\end{itemize}

\end{enumerate}
\end{definition}

\begin{example} Let $\Pi$ be a DFLP$^{\cal FA}$ program that consists of the DFLP$^{\cal FA}$ rules:
\[
\begin{array}{lcl}
a(1,1):0.8 \;\; \vee \;\; a(1,2):0.4 & \leftarrow &  \\
a(2, 1):0.3 \;\; \vee \;\; a(2,2):0.9 & \leftarrow &  \\
\end{array}
\]
\[
\begin{array}{l}
r:  \Gamma   \leftarrow not \; \Gamma, \: min_F \{ Y : U \: | \: a(X, Y): U  \} \leq 1 \: : \: 0.4 \\
\end{array}
\]
The ground instantiation of $r$ is given by:
\[
\begin{array}{lcl}
r': \Gamma  && \leftarrow not \; \Gamma, \: min_F \{ \\ &&
\langle 1 : 0.8 \: | \: a(1, 1): 0.8 \rangle,\; \langle 2 : 0.4 \: | \: a(1, 2): 0.4 \rangle , \\&&
\langle 1 : 0.3 \: | \: a(2, 1): 0.3 \rangle, \; \langle 2 : 0.9 \: | \:  a(2, 2): 0.9 \rangle \\&&
\} \leq 1 \; : \; 0.4\\
\end{array}
\]
Let $I$ be a fuzzy interpretation of $\Pi$ that assign $0.4$ to $a(1,2)$, $0.9$ to $a(2,2)$, and assigns $0$ to the remaining atoms in ${\cal B_L}$. Thus the evaluation of the fuzzy aggregate atom, $min_F(S) \leq 1$ in $r'$ w.r.t. to $I$ is given as follows, where
\[
\begin{array}{lcl}
S = \{ &&  \langle 1 : 0.8 \: | \: a(1, 1): 0.8 \rangle,\; \langle 2 : 0.4 \: | \: a(1, 2): 0.4 \rangle , \\&&
\langle 1 : 0.3 \: | \: a(2, 1): 0.3 \rangle, \; \langle 2 : 0.9 \: | \:  a(2, 2): 0.9 \rangle \: \}
\end{array}
\]
and $S_I = \{ 2:0.4, \; 2:0.9 \}$. Therefore, \\ $min_F( \{ 2:0.4, \; 2:0.9 \} ) = (2, 0.4)$, and consequently, the fuzzy annotated fuzzy annotated aggregate atom $min_F(S) \leq 1 \;  : \; 0.4$ is not satisfied by $I$. This is because $min_F( \{ 2:0.4, \; 2:0.9 \} ) = (2, 0.4) \neq \bot$ and $2 \nleq 1$ although $0.4 \leq 0.4$. Let $I'$ be a fuzzy interpretation of $\Pi$ that assign $0.8$ to $a(1,1)$, $0.9$ to $a(2,2)$, and assigns $0$ to the remaining atoms in ${\cal B_L}$. Thus, $S_{I'} = \{ 1:0.8, 2:0.9 \}$ and $min_F (\{ 1:0.8, 2:0.9 \}) = (1,0.8)$, hence
the fuzzy annotated fuzzy annotated aggregate atom $min_F(S) \leq 1 \;  : \; 0.4$ is satisfied by $I'$, since $1 \leq 1$ and $0.4 \leq 0.8$.

\label{ex:dice}
\end{example}
Let $A$ be a fuzzy annotated atom, $a:\mu$ or the negation of $a:\mu$, denoted by $not \; a:\mu$. Let $I_1, I_2$ be two fuzzy interpretations. Then, we say that $A$ is monotone if $\forall (I_1, I_2)$ such that $I_1 \leq I_2$, it is the case that if $I_1$ satisfies $A$ then $I_2$ also satisfies $A$. However, $A$ is antimonotone if $\forall (I_1, I_2)$ such that $I_1 \leq I_2$ it is the case that if $I_2$ satisfies $A$ then $I_1$ also satisfies $A$. But, if $A$ is not monotone or not antimonotone, then we say $A$ is nonmonotone. A fuzzy annotated atom or a fuzzy annotated fuzzy aggregate atom, $a:\mu$, or the negation of fuzzy annotated atom or the negation of a fuzzy annotated fuzzy aggregate atom, $not \; a:\mu$, can be monotone, antimonotone or nonmonotone, since their fuzzy annotations are allowed to be arbitrary functions. Moreover, fuzzy aggregate atoms by themselves can be monotone, antimonotone or nonmonotone.

\begin{definition} A fuzzy model for a DFLP$^{\cal FA}$ program, $\Pi$, is a fuzzy interpretation for $\Pi$ that satisfies $\Pi$. A fuzzy model $I$ for $\Pi$ is $\leq$--minimal iff there does not exist a fuzzy model $I'$ for $\Pi$ such that $I' < I$.
\end{definition}

\begin{example} It can easily verified that the fuzzy interpretation, $I$, for DFLP$^{\cal FA}$ program, $\Pi$, described in Example (\ref{ex:dice}), is a minimal fuzzy model for $\Pi$. However, the fuzzy interpretation, $I'$, for $\Pi$, described in Example (\ref{ex:dice}), is not a fuzzy model for $\Pi$.
\end{example}

\begin{definition} Let $\Pi$ be a ground DFLP$^{\cal FA}$ program, $r$ be a DFLP$^{\cal FA}$ rule in $\Pi$, and $I$ be a fuzzy interpretation for $\Pi$. Let $I \models body(r)$ denotes $I$ satisfies $body(r)$. Then, the fuzzy reduct, $\Pi^I$, of $\Pi$ w.r.t. $I$ is a ground DFLP$^{\cal FA}$ program $\Pi^I$ where
\[
\Pi^I = \{ head(r) \leftarrow body(r) \: \: | \: \: r \in \Pi \: \wedge \: I \models body(r)\}
\]
\end{definition}

\begin{definition} A fuzzy interpretation, $I$, of a ground DFLP$^{\cal FA}$ program, $\Pi$, is a fuzzy answer set for $\Pi$ if $I$ is $\leq$-minimal fuzzy model for $\Pi^I$.
\end{definition}
Observe that the definitions of the fuzzy reduct and the fuzzy answer sets for DFLP$^{\cal FA}$ programs are generalizations of the fuzzy reduct and the fuzzy answer sets of the original DFLP programs described in \cite{Saad_DFLP}.

\begin{example} It can be easily verified that the DFLP$^{\cal FA}$ program described in Example (\ref{ex:dice}) has three fuzzy answer sets $I_1$, $I_2$, and $I_3$ presented below, where atoms in $\cal B_L$ that are not appearing in $I_1$, $I_2$, and $I_3$ are assumed to be assigned the fuzzy annotation $0.0$.
\[
\begin{array}{c}
I_1 = \{a(1,1):0.8, a(2, 1):0.3 \} \\
I_2 = \{a(1,2):0.4, a(2, 1):0.3 \}  \\
I_3 = \{a(1,2):0.4, a(2,2):0.9\}
\end{array}
\]
\end{example}

\begin{example} The DFLP$^{\cal FA}$ program representation of the fuzzy company control problem, $\Pi$, described in Example (\ref{ex:main}) has one fuzzy answer set, $I$, which, after omitting the facts and assuming atoms in $\cal B_L$ that do not appear in $I$ are assigned the annotation $0.0$, is
\[
\begin{array}{lcl}
I = \{ && \\ controlStk(a, a, b, 40) & : & 0.7, \\ controlStk(a, a, c, 40) & : & 0.9, \\
controlStk(b, b, c, 20) & : & 0.8, \\ controlStk(c, c, b, 20)& : & 0.6 \}.
\end{array}
\]
The fuzzy answer set, $I$, implies that no company {\em fuzzy} controls another company.
\end{example}

\section{DFLP$^{\cal FA}$ Semantics Properties}

In this section we study the semantics properties of DFLP$^{\cal FA}$  programs and its relationship to the original fuzzy answer set semantics of disjunctive fuzzy logic programs, denoted by DFLP \cite{Saad_DFLP}; the classical answer set semantics of classical disjunctive logic programs with classical aggregates, denoted by DLP$^{\cal A}$ \cite{Recur-aggr}; and the original classical answer set semantics of classical disjunctive logic programs, denoted by DLP \cite{Gelfond_B}.

\begin{theorem} Let $\Pi$ be a DFLP$^{\cal FA}$ program. The fuzzy answer sets for $\Pi$ are $\leq$--minimal fuzzy models for $\Pi$.
\end{theorem}
The following theorem shows that the fuzzy answer set semantics of DFLP$^{\cal FA}$ subsumes and generalizes the fuzzy answer set semantics of DFLP \cite{Saad_DFLP}, which are DFLP$^{\cal FA}$ programs without fuzzy aggregates atoms and with only monotone fuzzy annotation functions.

\begin{theorem} Let $\Pi$ be a DFLP program and $I$ be a fuzzy interpretation. Then, $I$ is a fuzzy answer set for $\Pi$ iff $I$ is a fuzzy answer set for $\Pi$ w.r.t. the fuzzy answer set semantics of \cite{Saad_DFLP}.
\end{theorem}

Now we show that the fuzzy answer set semantics of DFLP$^{\cal FA}$ programs naturally subsumes and generalizes the classical answer set semantics of the classical disjunctive logic programs with the classical aggregates, DLP$^ {\cal A}$ \cite{Recur-aggr}, which consequently naturally subsumes the classical answer set semantics of the original classical disjunctive logic programs, DLP \cite{Gelfond_B}.

Any DLP$^{\cal A}$ program, $\Pi$, is represented as a DFLP$^{\cal FA}$ program, $\Pi'$, where each DLP$^{\cal A}$ rule in $\Pi$ of the form
\[
a_1 \; \vee \ldots \vee \; a_k \leftarrow a_{k+1}, \ldots, a_m, not\; a_{m+1},\ldots, not \;a_{n}
\]
is represented, in $\Pi'$, as a DFLP$^{\cal FA}$ rule of the form
\[
\begin{array}{r}
a_1:1 \; \vee \ldots \vee \; a_k:1 \leftarrow a_{k+1}:1, \ldots, a_m:1, \\ not\; a_{m+1}:1,\ldots, not \;a_{n}:1
\end{array}
\]
where $a_1, \ldots, a_k$ are atoms and $a_{k+1},\ldots, a_n$ are atoms or fuzzy aggregate atoms whose fuzzy aggregates contain fuzzy sets that involve conjunctions of fuzzy annotated atoms with the fuzzy annotation $1$, where $1$ represents the truth value \emph{true}. We call this class of DFLP$^{\cal FA}$ programs as DFLP$_1^{\cal FA}$. Any DLP program is represented as a DFLP$_1^{\cal FA}$ program by the same way as DLP$^{\cal A}$ except that DLP disallows classical aggregate atoms. The following results show that DFLP$_1^{\cal FA}$ programs subsume both DLP$^{\cal A}$ and DLP programs.

\begin{theorem} Let $\Pi'$ be a DFLP$_1 ^{\cal FA}$ program equivalent to a DLP$^{\cal A}$ program $\Pi$. Then, $I'$ is a fuzzy answer set for $\Pi'$ iff $I$ is a classical answer set for $\Pi$, where $I'(a) = 1$ iff $a \in I$ and $I'(b) = 0$ iff $b \in {\cal B_L} - I$.
\label{thm:DLP2DHPP}
\end{theorem}

\begin{proposition} Let $\Pi'$ be a DFLP$_1 ^{\cal FA}$ program equivalent to a DLP program $\Pi$. Then, $I'$ is a fuzzy answer set for $\Pi'$ iff $I$ is a classical answer set for $\Pi$, where $I'(a) = 1$ iff $a \in I$ and $I'(b) = 0$ iff $b \in {\cal B_L} - I$.
\end{proposition}

\section{Conclusions and Related Work}

We presented the syntax and semantics of the fuzzy aggregates disjunctive fuzzy logic programs, DFLP$^{\cal FA}$, that extends the original disjunctive fuzzy logic programs, DFLP \cite{Saad_DFLP}, with arbitrary fuzzy annotation functions and with arbitrary fuzzy aggregates. We introduced the fuzzy answer set semantics of DFLP$^{\cal FA}$ programs with arbitrary fuzzy aggregates including monotone, antimonotone, and nonmonotone fuzzy aggregates. We have shown that the fuzzy answer set semantics of DFLP$^{\cal FA}$  subsumes and generalizes the fuzzy answer set semantics of the original DFLP \cite{Saad_DFLP}. In addition, we proved that the fuzzy answer sets of DFLP$^{\cal FA}$ are minimal fuzzy models and consequently incomparable, which is an important property for nonmonotonic fuzzy reasoning. We have shown that the fuzzy answer set semantics of DFLP$^{\cal FA}$ subsumes and generalizes the classical answer set semantics of both the classical aggregates classical disjunctive logic programs and the original classical disjunctive logic programs. To the best of our knowledge, this development is the first to consider fuzzy aggregates in fuzzy logical reasoning in general and in fuzzy answer set programming in particular. However, classical aggregates were extensively investigated in classical answer set programming \cite{Recur-aggr,Smodels-Weight,WFM-Pref,Dis-mono-aggr,Ferraris,FOL-aggr,Pelov}. A comprehensive comparisons among these approaches to classical aggregates in classical answer set programming \cite{Recur-aggr,Smodels-Weight,WFM-Pref,Dis-mono-aggr,Ferraris,FOL-aggr,Pelov} in general and between these approaches and DLP$^{\cal A}$ in particular is found in \cite{Recur-aggr}.

\bibliographystyle{named}
\bibliography{Saad13FAFASP}

\begin{thebibliography}{}

\bibitem[\protect\citeauthoryear{Faber \bgroup \em et al.\egroup
  }{2010}]{Recur-aggr}
W.~Faber, N.~Leone, and G.~Pfeifer.
\newblock Semantics and complexity of recursive aggregates in answer set
  programming.
\newblock {\em Artificial Intelligence}, 2010.

\bibitem[\protect\citeauthoryear{Ferraris and Lifschitz}{2005}]{Ferraris}
P.~Ferraris and V.~Lifschitz.
\newblock Weight constraints as nested expressions.
\newblock {\em TPLP}, 5:45–--74, 2005.

\bibitem[\protect\citeauthoryear{Ferraris and Lifschitz}{2010}]{FOL-aggr}
P.~Ferraris and V.~Lifschitz.
\newblock On the stable model semantics of first-order formulas with
  aggregates.
\newblock In {\em Nonmonotonic Reasoning}, 2010.

\bibitem[\protect\citeauthoryear{Gelfond and Lifschitz}{1991}]{Gelfond_B}
M.~Gelfond and V.~Lifschitz.
\newblock Classical negation in logic programs and disjunctive databases.
\newblock {\em New Generation Computing}, 9(3-4):363--385, 1991.

\bibitem[\protect\citeauthoryear{Niemel\"{a} and Simons}{2001}]{Smodels-Weight}
I.~Niemel\"{a} and P.~Simons.
\newblock Extending the smodels system with cardinality and weight constraints.
\newblock In {\em Logic-Based AI}, 2001.

\bibitem[\protect\citeauthoryear{Pelov and Truszczynski}{2004}]{Dis-mono-aggr}
N.~Pelov and M.~Truszczynski.
\newblock Semantics of disjunctive programs with monotone aggregates - an
  operator-based approach.
\newblock In {\em NMR}, 2004.

\bibitem[\protect\citeauthoryear{Pelov \bgroup \em et al.\egroup
  }{2007}]{WFM-Pref}
N.~Pelov, M.~Denecker, and M.~Bruynooghe.
\newblock Well-fouded and stable semantics of logic programs with aggregates.
\newblock {\em TPLP}, 7:355–--375, 2007.

\bibitem[\protect\citeauthoryear{Pelov}{2004}]{Pelov}
N.~Pelov.
\newblock {\em Semantics of logic programs with aggregates}.
\newblock PhD thesis, Katholieke Universiteit Leuven, Leuven, Belgium, 2004.

\bibitem[\protect\citeauthoryear{Saad \bgroup \em et al.\egroup
  }{2009}]{Saad_IFSA}
E.~Saad, S.~Elmorsy, M.~Gaber, and Y.~Hassan.
\newblock Reasoning about actions in fuzzy environment.
\newblock In {\em World Congress of the International Fuzzy Systems
  Association/European society for Fuzzy Logic and Technology
  (IFSA/EUSFLAT-09)}, 2009.

\bibitem[\protect\citeauthoryear{Saad}{2009}]{Saad_EFLP}
E.~Saad.
\newblock Extended fuzzy logic programs with fuzzy answer set semantics.
\newblock In {\em 3rd International Conference on Scalable Uncertainty
  Management}, 2009.

\bibitem[\protect\citeauthoryear{Saad}{2010}]{Saad_DFLP}
E.~Saad.
\newblock Disjunctive fuzzy logic programs with fuzzy answer set semantics.
\newblock In {\em 4rd International Conference on Scalable Uncertainty
  Management}, 2010.

\bibitem[\protect\citeauthoryear{Subrahmanian}{1994}]{Subrahmanian_B}
V.S. Subrahmanian.
\newblock Amalgamating knowledge bases.
\newblock {\em ACM TDS}, 19(2):291--331, 1994.

\end{thebibliography}

\end{document}